\documentclass[lettersize,journal]{IEEEtran}
\usepackage{amsmath,amsfonts}
\usepackage{algorithmic}
\usepackage{algorithm}
\usepackage{array}
\usepackage[caption=false,font=normalsize,labelfont=sf,textfont=sf]{subfig}
\usepackage{textcomp}
\usepackage{stfloats}
\usepackage{url}
\usepackage{verbatim}
\usepackage{graphicx}
\usepackage{cite}
\usepackage{authblk}

\begin{document}

\title{Procedural Generation of Grain Orientations using the Wave Function Collapse Algorithm}

\author{Grace Magny-Fokam, Dylan Madisetti, and Jaafar El-Awady
\thanks{G. Magny-Fokam is with Chesapeake Math and IT Academy, Upper Marlboro, MD 20772 USA (e-mail: gmagnyfokam@cmitsouth.org).}
\thanks{D. Madisetti and J. El-Awady are with the Hopkins Extreme Materials Institute, Johns Hopkins University, Baltimore, MD 21218 USA (e-mail: madisetti@jhu.edu; jelawady@jhu.edu).}}


\maketitle

\begin{abstract}
Statistics of grain sizes and orientations in metals correlate to the material's mechanical properties. Reproducing representative volume elements for further analysis of deformation and failure in metals, like 316L stainless steel, is particularly important due to their wide use in manufacturing goods today. Two approaches, initially created for video games, were considered for the procedural generation of representative grain microstructures. The first is the Wave Function Collapse (WFC) algorithm, and the second is constraint propagation and probabilistic inference through Markov Junior, a free and open-source software. This study aimed to investigate these two algorithms' effectiveness in using reference electron backscatter diffraction (EBSD) maps and recreating a statistically similar one that could be used in further research. It utilized two stainless steel EBSD maps as references to test both algorithms. First, the WFC algorithm was too constricting and, thus, incapable of producing images that resembled EBSDs. The second, MarkovJunior, was much more effective in creating a Voronoi tessellation that could be used to create an EBSD map in Python. When comparing the results between the reference and the generated EBSD, we discovered that the orientation and volume fractions were extremely similar. With the study, it was concluded that MarkovJunior is an effective machine learning tool that can reproduce representative grain microstructures.
\end{abstract}


\section{Introduction}
\IEEEPARstart{G}{rain} orientation represents the statistics of the relative orientation of individual crystals in a material. In a material, the direction of each orientation can be represented in colorful electron backscatter diffraction maps, with each color representing a different orientation. Studying grain orientations from EBSD maps allows us to predict a material’s behavior during plastic deformation. Experimentally taking EBSD images is time intensive and expensive. One could build upon it using procedural generation to produce many different representative grain orientation maps using only one input EBSD image. Procedural Generation (PCG) is a process by which computer systems take in specific patterns and create new, similar outputs from those patterns. 

\begin{figure}[h]
\centering
\includegraphics[width=2.5in]{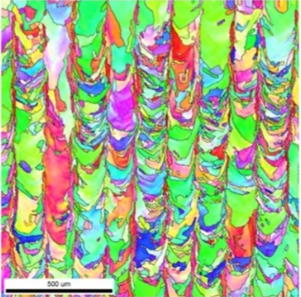}
\caption{Research \cite{ref6} on using PCG to simulate microstructures shows us that it’s possible to replicate EBSDs using PCG.\cite{ref1,ref6}}
\label{fig_1}
\end{figure}

\begin{figure}[h]
\centering
\includegraphics[width=2.5in]{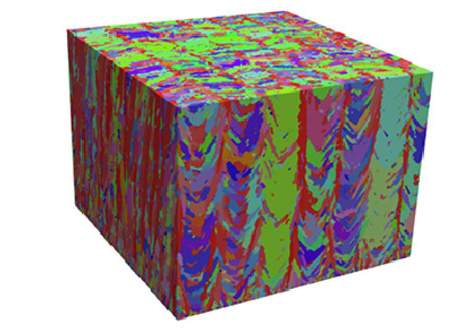}
\caption{Isometric view of EBSD map made from CAFE model \cite{ref6}}
\label{fig_17}
\end{figure}

\begin{figure}[h]
\centering
\includegraphics[width=2.5in]{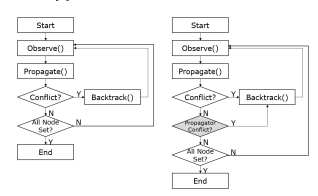}
\caption{Steps in general learning-based procedural generation \cite{ref5}}
\label{fig_2}
\end{figure}

Procedural Generation with the WFC algorithm could be used to input a simple stainless steel EBSD image map and to produce a statistically similar image based on the rules extracted from the input image. The WFC can generate bitmaps that are locally similar to the input bitmap \cite{ref5}. How the algorithm works is it analyzes the adjacency rules of an image, then makes guesses of what the surrounding area could look like based on those rules. It continues to do so until it sees a conflict, in which case it backtracks and repeats the process until there are no more conflicts \cite{ref5}. At this point, the run cycle is complete and an output image is produced. Figure \ref{fig_2} illustrates the steps the algorithm takes.

\newpage
\section{Literature Review}
\subsection{Granular Microstrutures and EBSDs}
To understand how one can accurately reproduce EBSD maps, we must first understand what they are and how they are created. An electron backscatter diffraction map is created when a scanning electron microscope (SEM) shoots an electron beam across the surface of a crystalline surface \cite{ref10}. Doing this reveals not only the orientation mapping of the material, but also the grain size and boundaries, the texture, and any signs of strain \cite{ref11}. 

The entire process of creating an EBSD is as short as it sounds, occurring in only a fraction of a second. How it works on a granular level is that first, the beam collects any patterns it recognizes in the material. Then, the EBSD camera extracts lines to determine orientations and map diffraction patterns. Finally, it matches orientations with a color legend. The specimen is broken up in a grid and this process repeats at each grid point until the figure map is complete \cite{ref11}.

Machine learning, especially computer vision, is quickly becoming commonplace in the field of materials science. Traditionally, humans are the ones who decide what to look for in material microstructure research and which methods they should employ to do so. However, in more recent studies, computer vision software has proven to be an adequate method for extracting visual information from polycrystalline microstructures \cite{ref12}. Prior research has shown that neural networks can be trained to find common themes between image representations \cite{ref12}, classify large amounts of EBSD patterns without human intervention \cite{ref13}, and accurately separate phases in a material by crystal symmetry \cite{ref14}. 

Despite these capabilities, due to the unique nature of EBSD maps, it's especially difficult and computationally intensive to train models to interpret EBSDs and reproduce them without the new microstructure losing its form. This could be primarily due to insufficient research and a lack of new techniques, which is why we brainstormed manners to resolve the issue. The most effective ideation was to use Voronoi tessellations.

A Voronoi tessellation is a diagram or pattern that divides a plane into separate regions. Traditionally, the pattern is created by scattering points on a plane, and then subdividing the plane into cells around each point. Each cell contains the portion of the plane that is closest to the corresponding point. However, in this case, what's being used is a centroidal Voronoi tessellation. In these tessellations, each Voronoi cell grows from a point or centroid (which is a cell's center of mass). The outward growth of each cell from its centroid is restricted to a certain mass by a given density function. Within the context of EBSDs, using a neural network to determine and notate the centroids of each orientation in a given EBSD map would allow the final output to retain as much similarity to the original as possible. Similar studies have shown the effectiveness of an intricate Voronoi tessellation in modeling granular geometry with significant precision and accuracy \cite{ref15}.

\subsection{Procedural Generation}

While procedural generation is most often used in game development and similar mediums, what's notable about that is vastly applicable is its ability to balance organization and randomness to achieve the most desirable outcomes. In a form of organized chaos, PCG follows any and all specific parameters given by the user and yet offers enough creativity to produce an intentionally random result. By harnessing this irregularity, under the right circumstances, PCG could be used to reproduce realistic EBSD maps, proving that it's possible to transplant this game development technique into the world of materials research.

\begin{figure}[h]
\centering
\includegraphics[width=2.5in]{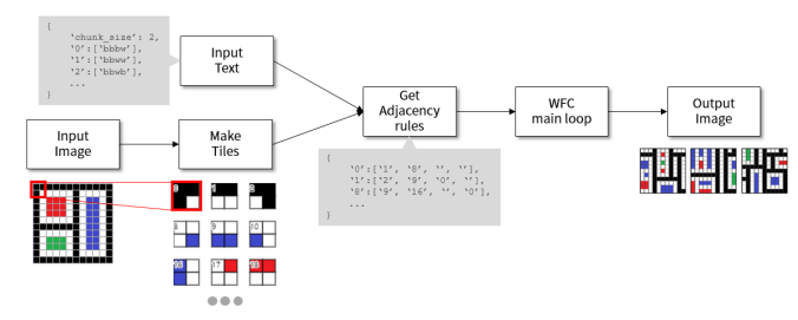}
\caption{Steps outlining the WFC's procedural generation algorithm\cite{ref5}}
\label{fig_3}
\end{figure}

\section{Procedural Generation Methods}
\subsection{Wave Function Collapse Algorithm}

First, the Python implementation of the WFC was used to test and train the model to first recognize, and then reproduce, an EBSD map \cite{ref4}. The algorithm utilizes texture synthesis to analyze visual patterns on a 2-D plane and synthesize an entirely different sample using the same texture. The problem, however, with texture synthesis, is that some assumptions are made when generating the sample, one being that the texels of the plane are known. \cite{ref8}
Another more important one is that the produced sample will fall into the two categories that encompass most textures: regular and stochastic. Real-world patterns such as EBSD maps tend not to fall into either of these categories \cite{ref8}. This is why we considered the aforementioned approach, utilizing a multi-parametric model that can synthesize a wider range of textures, including variable ones such as crystalline microstructures.

The selected implementation \cite{ref4} is an open-source software that was created and is primarily used in the creation of 2-D video game cartography like \textit{Super Mario Bros} \cite{ref9}. It had various parameters for optimization, ranging from location heuristics to backtracking propagation to the number of attempts before failure. 

\begin{figure}[h]
\centering
\includegraphics[width=2.5in]{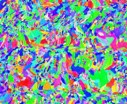}
\caption{316L Stainless Steel Reference EBSD \cite{ref7}}
\label{fig_6}
\end{figure}

By using Fig.\ref{fig_6} as a reference input, we did some parameter tuning in order to optimize the algorithm's ability to produce structures that look like granular orientations rather than cartoon terrains on which they were trained.  Two of the most critical parameters were the tile size on which the WFC iterates, and the pattern width. Tile size would determine the texels drawn from the reference, with 1 being the size of a single pixel. Pattern width had a bearing on the level of specificity during the propagation period. In this case, a higher pattern width meant a longer rendering period but with a supposedly more detailed and lifelike result. Manipulating those two while limiting the number of colors in the input image to 8 (for the sake of time efficiency during the rendering phase) created the opportunity to test the algorithm’s efficacy in producing statistically accurate outputs from a steel EBSD image. 

\begin{figure}[h]
\centering
\includegraphics[width=2.5in]{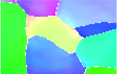}
\caption{Deformed Iron EBSD map \cite{ref2}}
\label{fig_4}
\end{figure}

\begin{figure}[h]
\centering
\includegraphics[width=2.5in]{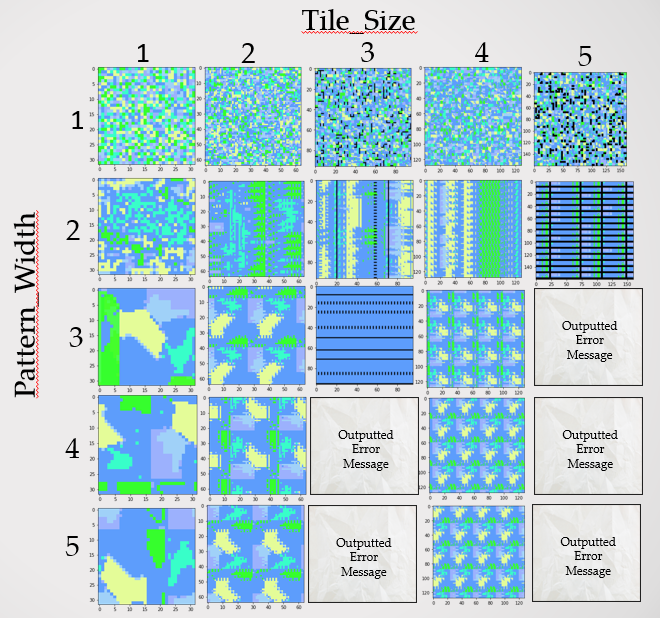}
\caption{Grid of WFC algorithm results}
\label{fig_5}
\end{figure}

At least 25 trials were run wherein parameters, namely tile size and pattern width, were manipulated on a scale of 1 to 5. For the sake of consistency, the same reference image (Figure  \ref{fig_4}) was used in each test, with each also being allotted equal rendering time. The results are depicted in Figure \ref{fig_5}.

The results that look closest to actual microstructures most often occur when the tile size is less than the pattern width. As can be perceived, the majority of the outputs from when pattern width was less than or equal to tile size aren't very representative, and thus could never be used in an analytical setting. The outputs from pattern width being high (over 3) and the tile size being less than 1 showed the algorithm would repeat patterns. This pattern recognition and repetition would be effective if it weren't for the fact that they're produced as an exact match of the original, whereas the goal was to have a statistically similar output that isn't entirely the same. Needless to say, the algorithm was also particularly dissimilar when it would produce sets of parallel black lines as a result of the pattern width and tile size being set to 3. Five of the 25 trials produced error messages due to conflicting parameters. 

Manipulating the parameters of this specific algorithmic implementation proved to be generally ineffective in accurately reproducing grain structures as it was too constrained or not constrained enough. Other tests run in conjunction with the preliminary trials also proved this to be true. Thus, it became necessary to develop a similar but undoubtedly different algorithm that had more malleable variables to eventually achieve the desired results. 

\subsection{MarkovJunior Algorithm}

Another implementation of the WFC was used in an algorithm called MarkovJunior \cite{ref3}, an open-source software that's based on Markov decision models and is written in C\# and XML. 

Markov decision processes, first ideated by Soviet mathematician Andrey Markov \cite{ref17}, are stochastic decision-making processes that use a probability-based mathematical framework wherein the results are partially random and partially controlled by a decision-maker. It's on this principle that Markov algorithms are built. They work in a similar fashion but instead rewrite strings over an alphabet based on a given set of rules. Below is a sample of the iteration of Markov algorithms.

\begin{figure}[h!]
\centering
\includegraphics[width=1in]{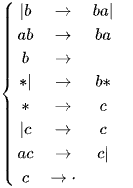}
\caption{Example of Markov algorithm scheme \cite{ref16}}
\label{fig_20}
\end{figure}

We used a Markov algorithm in collaboration with reinforcement learning to define and train the model to recognize the optimal behavior \cite{ref18} while maintaining a necessary degree of randomness. The following is the Bellman equation, a formula that's most optimal for solving stochastic optimal control problems.
\begin{equation}
    V(a)=\max _{0\leq c\leq a}\{u(c)+\beta V((1+r)(a-c))\}
\label{eqn:ExampleEquation}
\end{equation}

Both of these methods have proven to be successful in extracting meaningful information \cite{ref19}
from pre-existing EBSD maps, especially those made of metals like steel. The current concern, however, is to analyze its effectiveness in constraint propagation and replication.

Implementing this algorithm included extracting statistical information from a different stainless steel EBSD image and plugging them into MarkovJunior to produce another statistically similar EBSD.

This algorithm was also used to produce a Voronoi tessellation using the same data. When running MarkovJunior, statistics are extracted from the reference image using Python. From it, we gathered the total number of centroids, the volume fraction, and the orientation fraction from the reference. This information was then used to generate a Voronoi tessellation from the reference. Here’s what it looks like:

\begin{figure}[h!]
\centering
\includegraphics[width=2.5in]{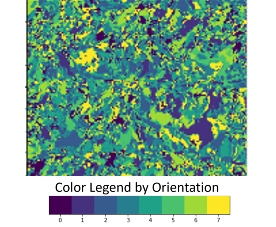}
\caption{Reference EBSD map (with color modifications)}
\label{fig_7}
\end{figure}

\begin{figure}[h!]
\centering
\includegraphics[width=2.5in]{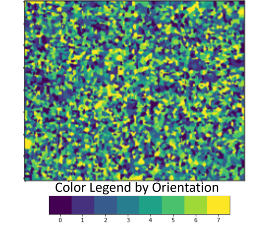}
\caption{Generated Voronoi tessellation reference}
\label{fig_8}
\end{figure}

2D and 3D visualizations of the Voronoi tessellation were also created. The 2D visualizations were created to be easier for scientists to analyze and compare, while the 3D gives a more realistic representation. 

\begin{figure}[h!]
\centering
\includegraphics[width=2.5in]{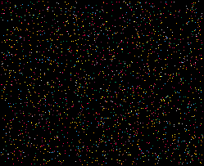}
\caption{Plotted Centroids for Voronoi tessellation growth}
\label{fig_9}
\end{figure}

\begin{figure}[h!]
\centering
\includegraphics[width=2.5in]{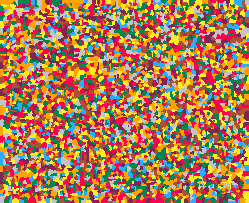}
\caption{Originally Generated Voronoi tessellation from MarkovJunior algorithm }
\label{fig_10}
\end{figure}

\begin{figure}[h!]
\centering
\includegraphics[width=2.5in]{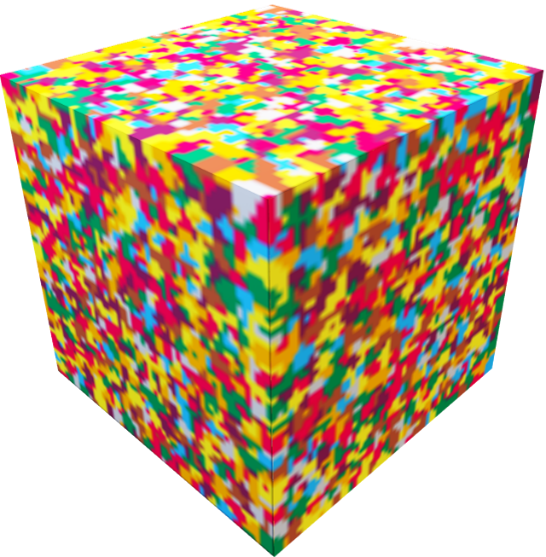}
\caption{3D Implementation of the original}
\label{fig_11}
\end{figure}

\begin{figure}[h!]
\centering
\begin{tabular}{|l|l|}
\hline
Reference Centroids      & Generated Centroids              \\ \hline
2527        & 2903                      \\ \hline
\end{tabular}
\caption{Grain Count Comparison}
\label{fig_16}
\end{figure}

\begin{figure}[h!]
\centering
\includegraphics[width=2.5in]{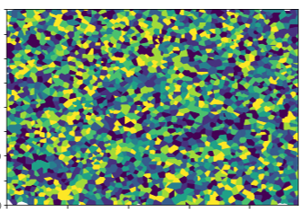}
\caption{Originally Generated Voronoi tessellation from MarkovJunior algorithm (with color modifications)}
\label{fig_12}
\end{figure}

\begin{figure}[h!]
\centering
\includegraphics[width=2.5in]{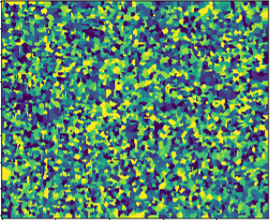}
\caption{Final Replicated EBSD map}
\label{fig_13}
\end{figure}

\newpage

\section{Analysis}
To check the accuracy of the PCG model, we compared the volume fraction, orientation fraction, and number of centroids between the reference and generated EBSDs.

\begin{figure}[h!]
\centering
\includegraphics[width=2.5in]{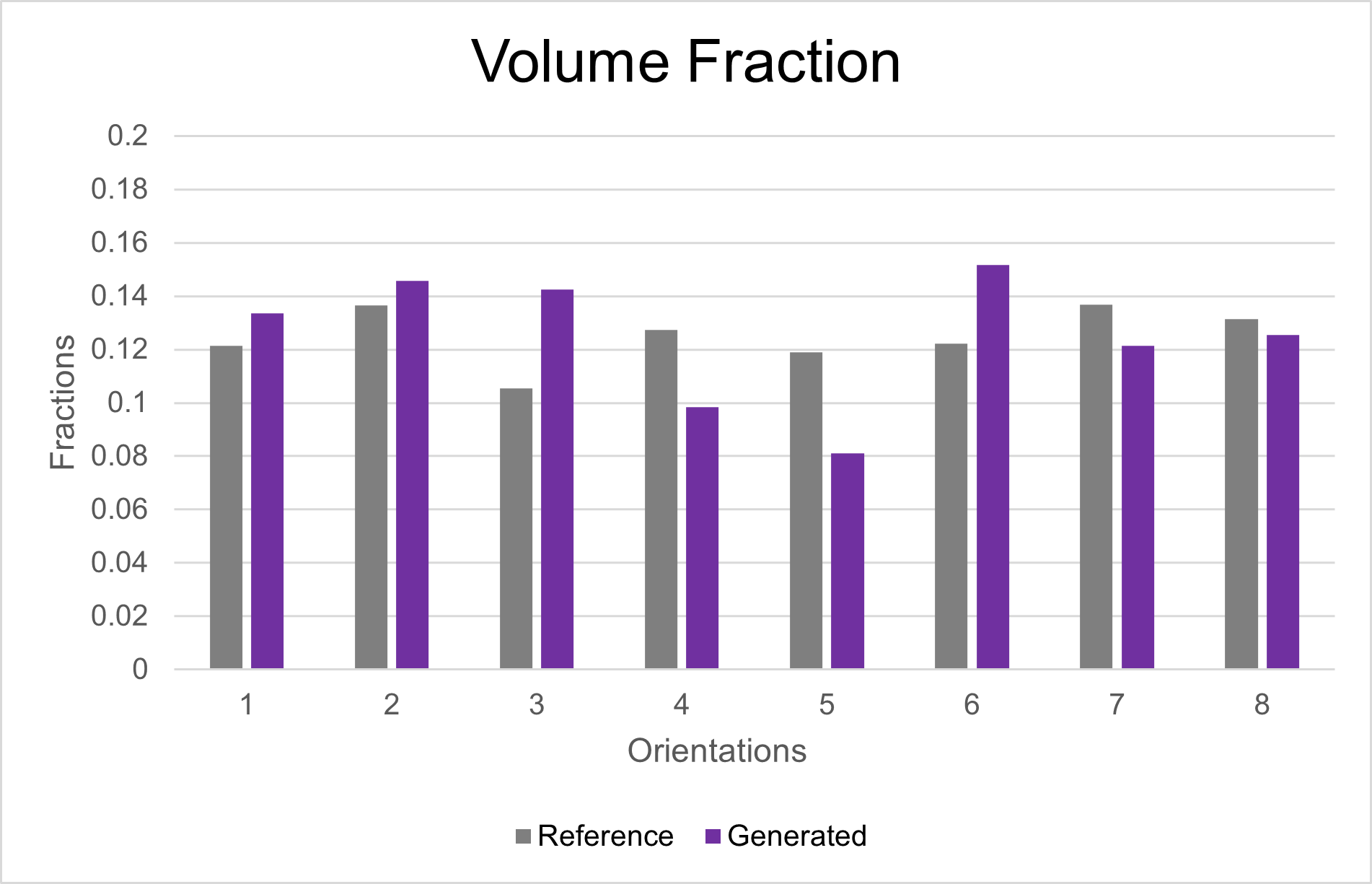}
\caption{Volume Fraction Comparison}
\label{fig_14}
\end{figure}

\begin{figure}[h!]
\centering
\includegraphics[width=2.5in]{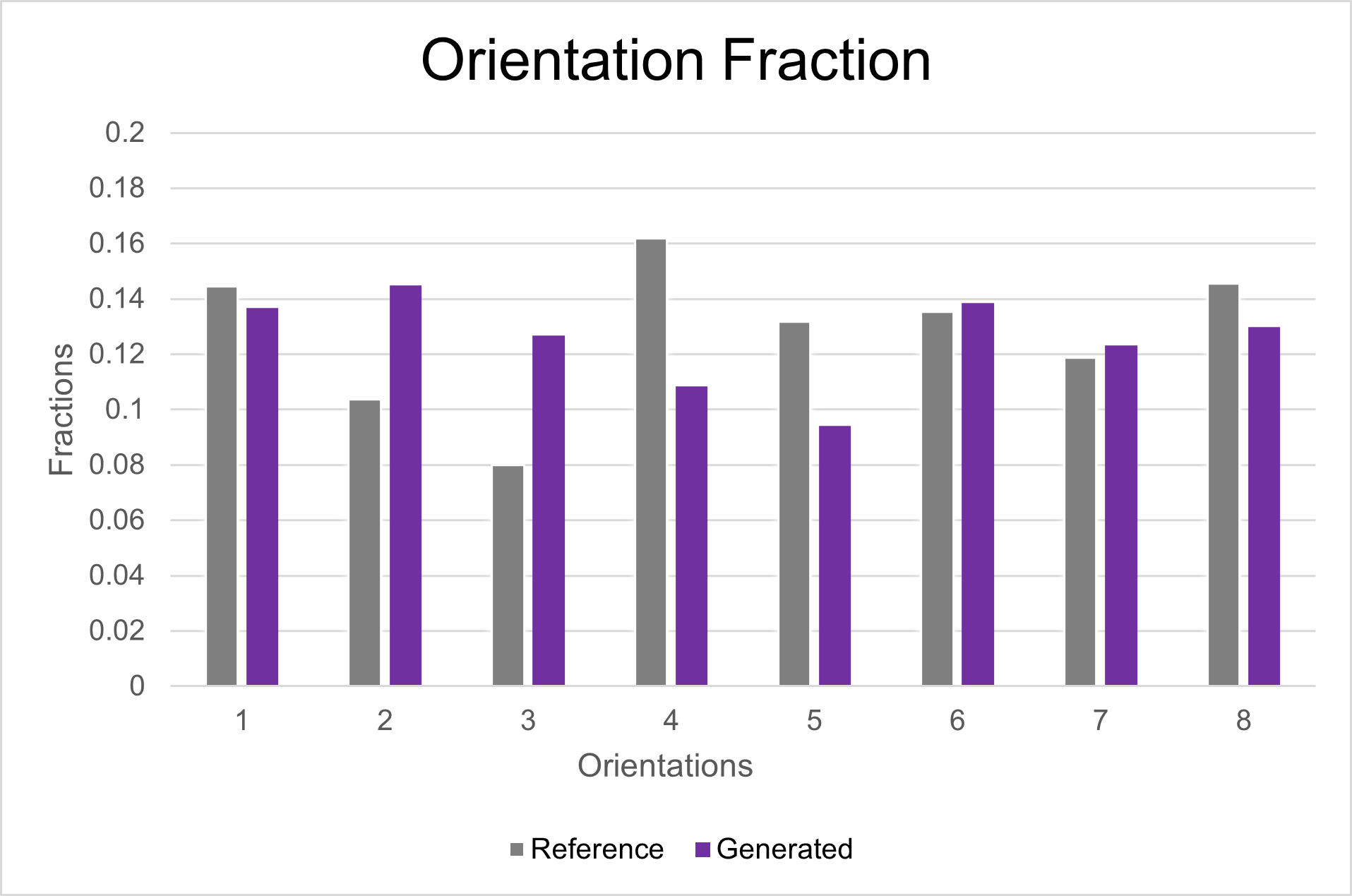}
\caption{Orientation Fraction Comparison}
\label{fig_15}
\end{figure}

As can be seen, the volume fractions, centroid count, and orientation fractions generated were close to but not exactly the same as those in the reference. While using the Wave Function Collapse (WFC) algorithm, we realized that it might be too constricting to reproduce EBSD maps effectively. Even after being modified several times, the outputs it was producing didn’t have the same structure as an EBSD. We switched to the MarkovJunior algorithm and found that with the help of a simple Python program, it could reproduce a Voronoi tessellation that could then be turned into an EBSD map. The generated EBSD maps were statistically similar to the original but different.

\section{Conclusions}
My set of experiments using the preliminary WFC algorithm showed promise but ultimately, didn’t produce the type of outputs I needed. The first WFC was too constraining for my project. However, the results of the WFC implementation helped point towards a less constricting model in MarkovJunior. In the latter one, data derived from EBSD images was able to successfully reproduce statistically similar ones. The broader implications of this study point towards one of the most pressing uses for reinforced 361L stainless steel, natural disaster preparedness and risk mitigation. By understanding how we can better study material microstructures with less human labor through AI, we inch closer to developing new ways to fortify these materials in ways that they are more resistant to inclement weather conditions and other unpredictable factors that govern our lives. Future work would include considering the adjacency of orientation types and using orientation non-categorically from EBSD maps.

\section{Acknowledgements}
This research was funded by the U.S. Department of Defense's Army Educational Outreach Program (AEOP) and supported by the Hopkins Extreme Materials Institute (HEMI) at Johns Hopkins University.

\vfill


\newpage
\begin{thebibliography}{1}
\bibliographystyle{IEEEtran}

\bibitem{ref1}
O. Andreau, I. Koutiri, P. Peyre, J. D. Penot, N. Saintier, E. Pessard, T. de Terris, C. Dupuy, T. Baudin, "Texture control of 316L parts by modulation of the melt pool morphology in selective laser melting,"{\emph{Journal of Materials Processing Technology}}, pp. 21-31, 2019, doi: 10.1016/j.jmatprotec.2018.08.049

\bibitem{ref2}
T. B. Britton and J. Hickey,“Deformed Iron EBSD data set”. Zenodo, Apr. 07, 2018, doi: 10.5281/zenodo.1214829

\bibitem{ref3}
Mxgmn, “GitHub - mxgmn/MarkovJunior: Probabilistic language based on pattern matching and constraint propagation, 153 examples,” GitHub. https://github.com/mxgmn/MarkovJunior

\bibitem{ref4}
Ikarth, “GitHub - ikarth/wfc\_2019f,” GitHub. https://github.com/ikarth/wfc\_2019f

\bibitem{ref5}
H. Kim, S-T. Lee, H. Lee, T. Hahn, and S-J. Kang,"Automatic Generation of Game Content using a Graph-based Wave Function Collapse Algorithm,"{\emph{2019 IEEE Conference on Games (CoG)}}, pp. 1-4, Aug 2019, doi: /10.1109/cig.2019.8848019

\bibitem{ref6}
K. Teferra and D. J. Rowenhorst, "Optimizing the cellular automata finite element model for additive manufacturing to simulate large microstructures,"{\emph{Acta Materialia}}, vol. 213, p. 116930, Jul 2021, doi: 10.1016/j.actamat.2021.116930

\bibitem{ref7}
C. Todaro, M. Easton, D. Qiu, J. M. BraGrain refinement of stainless steel in ultrasound-assisted additive manufacturing,"{\emph{Additive Manufacturing}}, vol. 37, p. 101632, Jan 2021, doi: 10.1016/j.addma.2020.101632

\bibitem{ref8}
A. A. Efros and T. Leung, “Texture synthesis by non-parametric sampling,” {\emph{Proceedings of the Seventh IEEE International Conference on Computer Vision}}, Sep 1999, doi: 10.1109/iccv.1999.790383.

\bibitem{ref9}
R. Hoeft and A. Nieznanska, “Empirical evaluation of procedural level generators for 2D platform games,”{\emph{ Dissertation}}, 2014.

\bibitem{ref10}
R. G. Mariano, A. Yau, J. T. McKeown, M. Kumar, and M. W. Kanan, “Comparing scanning electron microscope and transmission electron microscope grain mapping techniques applied to Well-Defined and Highly irregular nanoparticles,” {\emph{ACS Omega}}, vol. 5, no. 6, pp. 2791–2799, Feb. 2020, doi: 10.1021/acsomega.9b03505.

\bibitem{ref11}
J. R. Michael, “Introduction to EBSD and applications in Materials Science.,” Jul. 2018. [Online]. Available: https://www.osti.gov/servlets/purl/1570933

\bibitem{ref12}
E. A. Holm et al., “Overview: Computer Vision and Machine Learning for microstructural characterization and analysis,” {\emph{Metallurgical and Materials Transactions}}, vol. 51, no. 12, pp. 5985–5999, Sep. 2020, doi: 10.1007/s11661-020-06008-4.


\bibitem{ref13}
K. Kaufmann, H. Lane, X. Liu, and K. S. Vecchio, “Efficient few-shot machine learning for classification of EBSD patterns,” {\emph{Scientific Reports}}, vol. 11, no. 1, Apr. 2021, doi: 10.1038/s41598-021-87557-

\bibitem{ref14}
K. Kaufmann, C. Zhu, A. S. Rosengarten, D. Maryanovsky, H. Wang, and K. S. Vecchio, “Phase mapping in EBSD using convolutional neural networks,” {\emph{Microscopy and Microanalysis}}, vol. 26, no. 3, pp. 458–468, May 2020, doi: 5.10.1017/s1431927620001488.

\bibitem{ref15}
S. Ganesan, I. Javaheri, and V. Sundararaghavan, “Constrained Voronoi models for interpreting surface microstructural measurements,” {\emph{Mechanics of Materials}}, vol. 159, p. 103892, Aug. 2021, doi: 10.1016/j.mechmat.2021.103892.


\bibitem{ref16}
Caracciolo di Forino, A. String processing languages and generalized Markov algorithms. {\emph{In Symbol manipulation languages and techniques}}, D. G. Bobrow (Ed.), North-Holland Publ. Co., Amsterdam, the Netherlands, 1968, pp. 191–206.

\bibitem{ref17}
B. A. Kushner, “The Constructive Mathematics of A. A. Markov,” {\emph{The American Mathematical Monthly}}, vol. 113, no. 6, p. 559, Jun. 2006, doi: 10.2307/27641983.

\bibitem{ref18}
A. Bernstein and E. Burnaev, “Reinforcement learning in computer vision,” {\emph{Proceedings of the Tenth International Conference on Machine Vision (ICMV 2017)}}, vol. 10696, Apr. 2018, doi: 10.1117/12.2309945.


\bibitem{ref19}
T. M. Ostormujof, R. P. R. Purohit, S. Breumier, N. Gey, M. Salib, and L. Germain, “Deep Learning for automated phase segmentation in EBSD maps. A case study in Dual Phase steel microstructures,” {\emph{Materials Characterization}}, vol. 184, p. 111638, Feb. 2022, doi: 10.1016/j.matchar.2021.111638.


\end{thebibliography}
\end{document}